%% file: acl_latex.tex
\pdfoutput=1

\documentclass[11pt]{article}

\usepackage{acl}
\usepackage{hyperref}       
\usepackage{url}            
\usepackage{booktabs}       
\usepackage{amsfonts}       
\usepackage{nicefrac}       
\usepackage{microtype}      

\usepackage{times}
\usepackage{latexsym}
\usepackage{graphicx,wrapfig}
\usepackage{algorithm}
\usepackage{algorithmic}
\usepackage{enumitem}
\usepackage{amsthm} 
\usepackage{amsmath}
\usepackage{amssymb}
\usepackage{babel}
\usepackage{xcolor}
\usepackage{subfig}
\usepackage{multirow}

\usepackage[T1]{fontenc}

\usepackage[utf8]{inputenc}

\usepackage{microtype}

\title{Transfer Learning for Structured Pruning under Limited Task Data}

\author{Lucio Dery \\
  CMU\\
  \texttt{ldery@andrew.cmu.edu}
  \And
  David Grangier \\
  Apple \\
  \texttt{lastname@apple.com}
  \And
  Awni Hannun \\
  Apple \\ 
  \texttt{awni@apple.com}
}

\begin{document}
\maketitle
\input{sections/abstract}
\input{sections/introduction}
\input{sections/relatedwork}
\input{sections/methodology}
\input{sections/experimentaldetails}
\input{sections/resultsanddiscussion}
\newpage
\bibliography{acl_latex}
\input{sections/appendix}

\end{document}

%% file: sections/abstract.tex
\begin{abstract}
Large, pre-trained models are problematic to use in resource constrained applications. Fortunately, task-aware structured pruning methods offer a solution. These approaches reduce model size by dropping structural units like layers and attention heads in a manner that takes into account the end-task. However, these pruning algorithms require more task-specific data than is typically available. We propose a framework which combines structured pruning with transfer learning to reduce the need for task-specific data. Our empirical results answer questions such as: How should the two tasks be coupled? What parameters should be transferred? And, when during training should transfer learning be introduced? Leveraging these insights, we demonstrate that our framework results in pruned models with improved generalization over strong baselines.
\end{abstract}

%% file: sections/introduction.tex
\section{Introduction}
Large pre-trained language models have been successfully applied to a wide variety of application scenarios \citep{bommasani2021opportunities, anil2023palm}. However, not all applications can justify the cost of running such large models. E.g. an interactive, offline spellchecker for a phone has strong memory limits compared to a server-side chat model~\citep{dettmers2022llm}. Even server-side, the benefit/cost of large models depends on the application. This situation motivates research into structured model pruning algorithms. 

\begin{figure}
    \centering
    \includegraphics[width=0.49\textwidth]{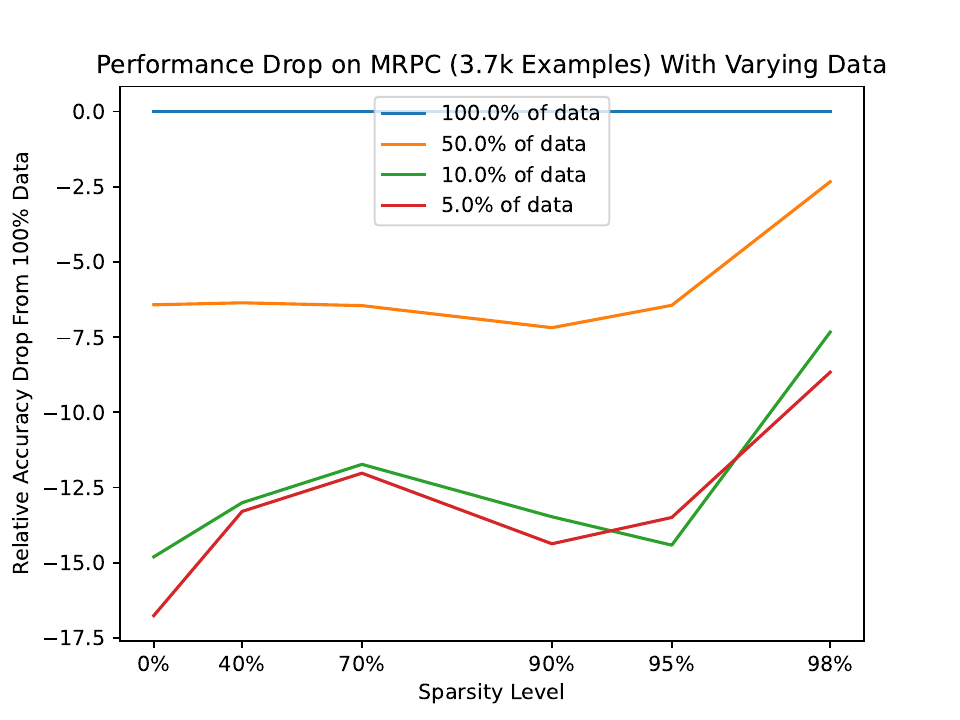}
    \caption{\label{fig:perf_drop_with_data} Accuracy degradation of CoFi \citep{xia2022structured} vs training data sizes. Sparsity level refers to the fraction of removed weights (excluding embeddings). Accuracy at 50\% data is stable across sparsity levels (except for 98\% sparsity) while more data-limited regimes (10\%--5\%) exhibit stronger sensitivity to the sparsity level.}
    \vspace{-3mm}
\end{figure}

Structured pruning algorithms generate smaller, faster and yet reasonably accurate sub-models from large pre-trained ones by removing components (beyond individual parameters) like convolutional channels, attention heads and whole layers. Several works over the years \citep{wang2019structured, sanh2020movement, xia2022structured} have been proposed to perform task-specific structured pruning. Unfortunately, to the best of our knowledge, all existing algorithms have been developed without consideration for the amount of training data available for the target task. Thus, as Figure \ref{fig:perf_drop_with_data} shows that, even state-of-the-art methods like CoFi \citep{xia2022structured}, do not gracefully handle scenarios with limited training data. We argue that the data-limited structured pruning setting is important since limited compute for inference and data scarcity for training tend to co-occur often in practice \cite{ahia2021low}.
A popular remedy to the limited data problem at fixed model size, is to leverage transfer learning \citep{caruana1997multitask, erhan2010does, dery2022aang} by introducing external data or extra tasks. In this work, we investigate transfer learning based remedies for structured pruning under limited data. Structured pruning algorithms need to jointly learn both model weights and structural variables (which layers, attention heads, etc. to prune) for the final size-reduced model \cite{wang2019structured, xia2022structured}. This added complexity makes deploying transfer learning in the structured pruning setting non-trivial and raises several questions. Do we only perform transfer learning for model weights or do we include structural variables too? How do we learn structural variables for the target task in a way that benefits from the presence of a transfer task? When is it best to introduce transfer learning so as to produce the most accurate pruned target model? 

This work aims to provide answers to the questions above. We propose a simple modification to existing structured pruning algorithms to allow for effective transfer of both structural variables and model parameters. Overall, our analyses allow us to provide prescriptions to researchers about what, how and when to transfer during structured pruning. Our effort results in significant improvements in generalization performance even at compression ratios as high as $50\times$.

%% file: sections/relatedwork.tex
\section{Background}
\paragraph{Unstructured Pruning} approaches sparsify models by zeroing out individual components of weight matrices \citep{frankle2018lottery, sanh2020movement}. The resulting sparse matrices reduce the memory overhead of the model but run-time gains cannot be realized unless on specialized hardware \citep{liu2018rethinking, ma2021non}. 
Over the years, many criteria for choosing which parameters to remove have been explored. Some approaches like magnitude pruning \cite{han2015learning} and Wanda \cite{sun2023simple} prune parameters based on either their magnitudes or the magnitude of their product with previous layer activations respectively. Other approaches like \cite{frankle2018lottery, sanh2020movement} use information about about much parameters have changed since initialization whilst others learn unstructured masks based using gradient descent \citep{Ramanujan_2020_CVPR}. \citet{ahia2021low} introduce the term \emph{the low-resource double-bind} for the challenge of compressing models in data limited regimes. Unlike us, they study magnitude pruning, which as mentioned, does not ordinarily lead to run-time gains. They also do not propose a remedy for the limited-data problem, which we do in this paper. 
\paragraph{Structured Pruning}
algorithms remove whole components  from pre-trained models such as attention heads \citep{michel2019sixteen, voita2019analyzing}, whole layers \citep{fan2019reducing}  or intermediate dimensions of fully connected layers \citep{wang2019structured} in order to produce faster, memory efficient sub-models without overly sacrificing downstream accuracy. Unlike unstructured pruning, there is no need for specialized hardware in order to realize the run-time speedups from compression. These approaches require optimizing over structural variables (to decide which model components to prune) and model weights (to adapt the final model to the disruption that results from removing whole components). Joint optimizations like these mean more variables to learn, resulting in the need for mode end-task data points. To the best of our knowledge, we are the first consider the challenge of structured pruning under limited data.
\paragraph{Other model compression approaches} Quantization methods \citep{polino2018model, dettmers2022llm} reduce model size by reducing the number of bits required to represent each weight. These methods are generally complementary to pruning approaches but only achieve maximum size reductions on the order of 2-4$\times$ before substantial model performance degradation. We are interested in achieving extreme compressions to the order of 50$\times$ reduction without significant loss in performance. Distilling directly to a target task has been shown to be a data-hungry process \citep{jiao2019tinybert}, often requiring a general distillation step (on abundant external data) to be able to achieve competitive performance with approaches modern structured pruning methods like CoFi \cite{xia2022structured}.
\paragraph{Multitask Transfer Learning} \citep{caruana1997multitask} is a common recipe for improving a models average performance on a desired end-task. When the end-task is data-limited, auxiliary tasks can be multi-tasked with the end-task \citep{dery2021auxiliary, dery2021should} to serve as proxy data. Previous work at the intersection of pruning and multitasking have only studied how to prune multi-task models \citep{garg2023structured, yang2023taskspecific}. Unlike these, our starting point is not a multitask model but a generalist pre-trained model like BERT \citep{devlin2018bert}.
Our work is interested in using multitasking in as much as it improves generalization of the pruned model with respect to the data-starved end-task only.

%% file: sections/methodology.tex
\section{Methodology}
\label{sec:methodology}
The goal of this paper is to improve the generalization of pruned models when the end-task is data-limited without sacrificing memory and run-time gains. We assume that we are given a structured pruning algorithm that jointly learns structural/masking variables (which we denote as $\{\mathbf{z}^{k}_{\mathrm{target}}\}$) and their corresponding parameters $\{\mathbf{\theta}_{\mathrm{target}}^{k}\}$ for the target end-task. $k$ indexes the set of $K$ structural variables being explored. We are primarily concerned with how to incorporate a transfer task by learning $\left[\{\mathbf{z}_{\mathrm{transfer}}^{k}\}, \{\mathbf{\theta}_{\mathrm{transfer}}^{k}\}\right]$ such that we enjoy improved generalization with the target task's final model. Since our focus is on the limited-data problem, we care less about a specific structured pruning algorithm and more about how to adapt any appropriate algorithm in the data starved setting. We therefore focus on building on top of a state-of-the-art structured pruning algorithm, CoFi \cite{xia2022structured} which we take as a representative algorithm. Whilst we describe CoFi below to provide sufficient background, for the rest of the paper, we will abstract away the details of the pruning algorithm and focus on the specifics of adapting transfer learning to this setting.

\subsection{CoFi}
CoFi (\textbf{Co}arse- and \textbf{Fi}ne-grained Pruning) is a mixed resolution structured pruning algorithm. Previous algorithms 
to prune transformer models~\cite{vaswani2017transformer} have focused on removing high level units like whole layers \citep{fan2019reducing} or finer grained modules like attention heads \cite{voita2019analyzing} and dimensions of fully connected layers \citep{wang2019structured} but not both types. CoFi 
introduces variables that account for pruning at multiple levels of granularity. \\
\textbf{Coarse Grain}: Each transformer layer consists of a multi-headed attention component that feeds into a fully connected two-layer non-linear perceptron  \cite{vaswani2017transformer}. CoFi introduces variables sets $\{z^{i}_{\mathrm{MHA}}\}_{i \in [N]}$ and $\{z^{i}_{\mathrm{FFN}}\}_{i \in [N]}$ for each of the model $N$ layers. $z^{i}_{\mathrm{MHA}}$ denotes the probability that the \textbf{whole} attention component of the $i$th layer is removed whilst  $z^{i}_{\mathrm{FFN}}$ is similarly defined for the fully connected component of the specified layer. CoFi also removes whole columns of the residual stream: $z^{\ell} \in \mathbb{R}^{d} \rightarrow \widehat{z}^{\ell} \in \mathbb{R}^{\hat{d}} \quad \forall \ell \in [N]$. For a BERT model, $d = 768$ is typically reduced to $d \approx 750$. \cite{xia2022structured} find that though relatively few columns are dropped, including columns as structural variables is important for producnig performant compressed models. \\
\textbf{Fine Grain}: Given a particular layer $i$, CoFi prunes subsets of the attention heads available. The variables $\{z^{i}_{j, \mathrm{head}}\}_{[j \in n_h]}$ represent the $j$th attention head in the $i$ layer which has $n_h$ total attention heads. A similar set of variables is defined for the fully connected units within a layer : $\{z^{i}_{j, \mathrm{fc}}\}_{[j \in n_f]}$ where the $i$th fully connected layer has $n_f$ units.

For the $jth$ attention head of the $i$th layer, the likelihood that this head is left unpruned is proportional to $z^{i}_{\mathrm{MHA}} \cdot z^{i}_{j, \mathrm{head}}$. This allows the algorithm to make coupled fine and coarse grained decisions that lead to improved results.
We collectively represent $\{\mathbf{z}\}$ as the set of all structural variables that are learned by CoFi. For a model with parameters $\theta$, $\{\mathbf{z}\}$ are learned by applying the reparameterisation trick on the hard concrete distribution \citep{louizos2018learning} and minimizing a joint loss wrt $\{\mathbf{z}, \theta\}$ that includes 
\begin{enumerate}
    \item distance from target size. CoFi follows \citet{wang2019structured} and adds a lagrangian term that penalizes deviations from the target sparsity.
    \item target task loss. Practitioners ultimately want a pruned model that generalizes well on their end-task. CoFi jointly optimizes the target task loss along with the pruning objective in order to produce performant pruned models.
    \item a distillation objective on the original large model. Following \citet{sanh2020movement}, CoFi jointly performs distillation and structured pruning by introducing a layer-wise distillation objective. 
\end{enumerate}
With these high level details in mind, we proceed to present our simple, transfer learning based modification to CoFi that leads to improved results in data-limited settings. 

\subsection{Transfer Learning for Structured Pruning under limited data}
\label{subsec:transfer_how}
Given a target task $\mathbf{T}$ with limited training data, we want to improve the final model generated by CoFi through leveraging additional training data from an auxiliary task $\mathbf{A}$. Let $\{\mathbf{z}_{\mathbf{T}}, \theta\}$ be the initial set of all structural variables and model parameters for the target task and $\{\widehat{\mathbf{z}}_{\mathbf{T}}, \widehat{\theta}\}^{\gamma}$ be final output of CoFi at a chosen sparsity level $\gamma$. $\widehat{\mathbf{z}}$ are binary variables $\widehat{z}_{i} \in \{0, 1\}$ which indicate whether component $i$ is dropped/masked out (0) or is retained (1). We would like a procedure that leverages the auxiliary task (with its own set of variables $\{\mathbf{z}_{\mathbf{A}}\}$ such that the generalization performance of the pruned model when using using data from $\mathbf{A}$ and $\mathbf{T}$ jointly improves upon using only data from  $\mathbf{T}$.  

There are several design questions that arise in this setting when thinking about how to effectively utilize $\mathbf{A}$. 
In the following sections, we discuss some of these pertinent questions and propose some reasonable choices which we will later experimentally validate.
\subsubsection{What criteria do we use to select the auxiliary task $\mathbf{A}$ ?}
\label{subsec:how_select_task}
The choice of auxiliary task, $\mathbf{A}$, is an important design decision that must be considered carefully. A poor choice could result in poor generalization performance (with respect to $\mathbf{T}$ ) of the pruned model instead of being helpful. To this end, inspired by existing literature, we propose two criteria for evaluating what auxiliary task to leverage:
\paragraph{(1) \textbf{resourcedness}:} Previous work on transfer learning for learning model parameters has demonstrated the benefits of leveraging large pools of data (which may possibly be unrelated to the eventual end-task) for pre-traing \citep{anil2023palm} or multi-tasking \citep{dery2021should}. We therefore have a strong prior that using data-rich auxiliary tasks might be helpful for also learning structural parameters even if they are unrelated to the end-task.
\paragraph{(2) \textbf{task-similarity}} Both theoretical \citep{baxter2000model, maurer2016benefit, dery2022aang} and empirical works \citep{gururangan2020don, dery2022aang} have shown that transfer learning works best when the auxiliary task is similar or related to the end-task. As a proxy for similarity, we consider auxiliary tasks that are from the same \emph{domain} as the end-task. 
\subsubsection{When should we introduce $\mathbf{A}$?}
\label{subsec:when_introduce_transfer}
Structured pruning approaches like CoFi usually perform a two stage process. In the first stage, they generate a pruned model at the desired sparsity level; this involves learning both $\{\widehat{\mathbf{z}}, \widehat{\theta}\}^{\gamma}_{\mathbf{T}}$. In the second stage, pruned model is then fine-tuned on the end-task by updating only $\widehat{\theta}^{\gamma}_{\mathbf{T}}$ keeping $\widehat{\mathbf{z}}^{\gamma}_{\mathbf{T}}$ fixed. The auxiliary task can be introduced in either or both of these stages. We explore following choices: 
\paragraph{Prune$(A) \rightarrow $ FT$(T)$:} We do structural pruning to learn both the weights and structure for a small model using \textbf{\emph{only}} the transfer task, $\mathbf{A}$: $\{\widehat{\mathbf{z}}, \widehat{\theta}\}^{\gamma}_{\mathbf{A}}$. We then fine-tune (FT) the pruned model on the target task (T) only to obtain $\widehat{\theta}^{\gamma}_{\mathbf{T}}$. Here, the target task is used only in the final fine-tuning stage and is not involved in learning the pruned model structure.
\paragraph{Prune$(T) \rightarrow $ FT$(A, T)$:} We learn both the weights and structure for a small model using the target task: $\{\widehat{\mathbf{z}}, \widehat{\theta}\}^{\gamma}_{\mathbf{T}}$. We share the pruned model parameters $\widehat{\theta}^{\gamma}$ and fine-tune on both (T) and (A). Here, the auxiliary task is used only in the final fine-tuning stage and is not involved in learning the pruned model structure.
\paragraph{Prune$(A, T) \rightarrow $ FT$(T)$:} We learn both the weights and structure for a small model using both the transfer and end-task: $\{\widehat{\mathbf{z}}_{\mathbf{T}}, \widehat{\mathbf{z}}_{\mathbf{A}}, \widehat{\theta}\}^{\gamma}$ are learned jointly (we will explore how in Section \ref{subsubsec:how_learn_joint}). We then fine-tune (FT) the pruned model weights on the target task (T) only.
\paragraph{Prune$(A, T) \rightarrow $ FT$(T, A)$:} We learn both the weights and structure for a small model using both the transfer and end-task: $\{\widehat{\mathbf{z}}_{\mathbf{T}}, \widehat{\mathbf{z}}_{\mathbf{A}}, \widehat{\theta}\}^{\gamma}$ are learned jointly (we will explore how in Section \ref{subsubsec:how_learn_joint}). We then fine-tune (FT) the pruned model weights on \textbf{both} the target and auxiliary tasks.
\subsubsection{How do we incorporate $\mathbf{A}$ when optimizing for $\{\widehat{\mathbf{z}}_{\mathbf{T}}, \widehat{\theta}\}^{\gamma}$}
\label{subsubsec:how_learn_joint}
When using the auxiliary task directly during pruning, there is the question of what the best way to jointly optimize  $\{\widehat{\mathbf{z}}_{\mathbf{T}}, \widehat{\mathbf{z}}_{\mathbf{A}}, \widehat{\theta}\}^{\gamma}$ such that we achieve improved generalization for the final pruned model with respect to $\mathbf{T}$. Note that we are assuming that the model parameter weights $\theta$ are shared between the two tasks but the structural variables are separate. This is because there are many more model parameters than structural variables $\|\theta\|_{0} \gg \|\mathbf{z}_{\mathbf{T}}\|_{0} + \|\mathbf{z}_{\mathbf{A}}\|_{0}$. And so introducing separate model weights for the auxiliary task presents a much more significant modelling overhead than introducing new structural variables.

We can explore different strategies for sharing variables across the two tasks such that the $\mathbf{T}$ benefits from $\mathbf{A}$.
\paragraph{Single mask multi-task}
learns a single set of structural $\{\mathbf{z}\}$ and  model $\{\mathbf{\theta}\}$ parameters that are shared between both tasks. This choice tightly couples the two tasks. Whilst this allows maximal sharing of information between the target and transfer task, poor choices of transfer tasks could cause this to perform worse than no transfer at all.
\paragraph{Multi-mask multi-task} learns distinct structural parameters $\{\mathbf{z}\}_{\mathbf{T}}$ and  $\{\mathbf{z}\}_{\mathbf{A}}$ for each task but a single set of model parameters $\{\mathbf{\theta}\}$ is shared between both tasks. There is no transfer of structural information and only the shared model parameters provide a coupling of the two tasks.
\paragraph{Our $\delta$-Formulation} aims to leverage strength from both alternatives. We propose this method where both tasks share a base set of structural variables $\{\mathbf{z}\}_{\mathrm{base}}$ but also have task specific addends such that: 
$\{\mathbf{z}\}_{\mathbf{T}} = \{\mathbf{z}_{\mathrm{base}} + \delta_{\mathbf{T}}\}$ and $\{\mathbf{z}\}_{\mathbf{A}} = \{\mathbf{z}_{\mathrm{base}} + \delta_{\mathbf{A}}\}$. We regularize $\delta_{\ast}$ to encourage sharing between tasks via $\mathbf{z}_{\mathrm{base}}$ whilst maintaining flexibility for task-specific modelling.

%% file: sections/experimentaldetails.tex
\section{Experimental Setup}
Our experimental framework is introduced to investigate the questions posed in the previous section.\\
\textbf{Datasets} ~ We consider 3 pairs of tasks. One pair of classification tasks are from the computer science domain tasks -- SCIIE \citep{luan2018multi} and ACL-ARC  \citep{jurgens2018measuring} with 3.2k and 3.7k training samples respectively. The second pair of tasks are biomedical domain tasks - RCT \citep{dernoncourt2017pubmed} (we artificially create a low-resource version of this task with 10k training samples) and CHEMPROT \citep{kringelum2016chemprot} which has 4.2k training samples. We use GLUE \citep{wang2018glue} tasks for our last pair: STSB and MRPC are sentence similarity and paraphrase detection tasks with 7k and 3.7k train examples respectively. For the GLUE tasks, we follow previous work \citep{jiao2019tinybert, wang2019structured, xia2022structured} and report results on the validation set. For Non-GLUE tasks, we report test set results.  Please see Appendix \ref{app:data_info} for more details about the tasks we investigate. \\
\textbf{Model Details} ~ Since we use CoFi \citep{xia2022structured} as our representative structured pruning algorithm, we use the same model configuration. We use the BERT$_{\mathrm{base}}$ \citep{devlin2018bert} which has $\sim 110$M parameters.  We explore pruned model sparsities in the set $\{40\%, 70\%, 90\%, 95\%, 98\% \}$. $\gamma$\% sparsity means that the model has been reduced to $(100 - \gamma)\% \times  110$M parameters. Similar to \cite{sanh2020movement} we also freeze the model embedding weights. See Appendix \ref{app:training_details} for details about training as well as hyper-parameter values. \\
\textbf{Training details} ~ We mostly follow the training recipe from CoFi with a few minor changes. CoFi assumes that \emph{one starts pruning after finetuning the full parent model on the target task} and so introduces a distillation loss as part of the pruning objective. In our case, we start directly from the pre-trained model without first fine-tuning on the target task. This is because of the risk of over-fitting due to the smaller target task size.  Due to this, we find that the distillation based losses from the original CoFi paper are unnecessary and we did not see significant performance differences with or without them. When multitasking, we explore a small set of weighting hyper-parameters $\{(1.0, 1.0)~,~(1.0, 2.0)~,~(2.0, 1.0)\}$ for any losses relating to the target and auxiliary tasks respectively. Table \ref{tab:hp_choices} has details of the hyper-parameters we cross validate against for all our experiments.

%% file: sections/resultsanddiscussion.tex
\section{Empirical Recommendations for practitioners}
\begin{figure}
  \begin{center}
    \includegraphics[scale=0.22]{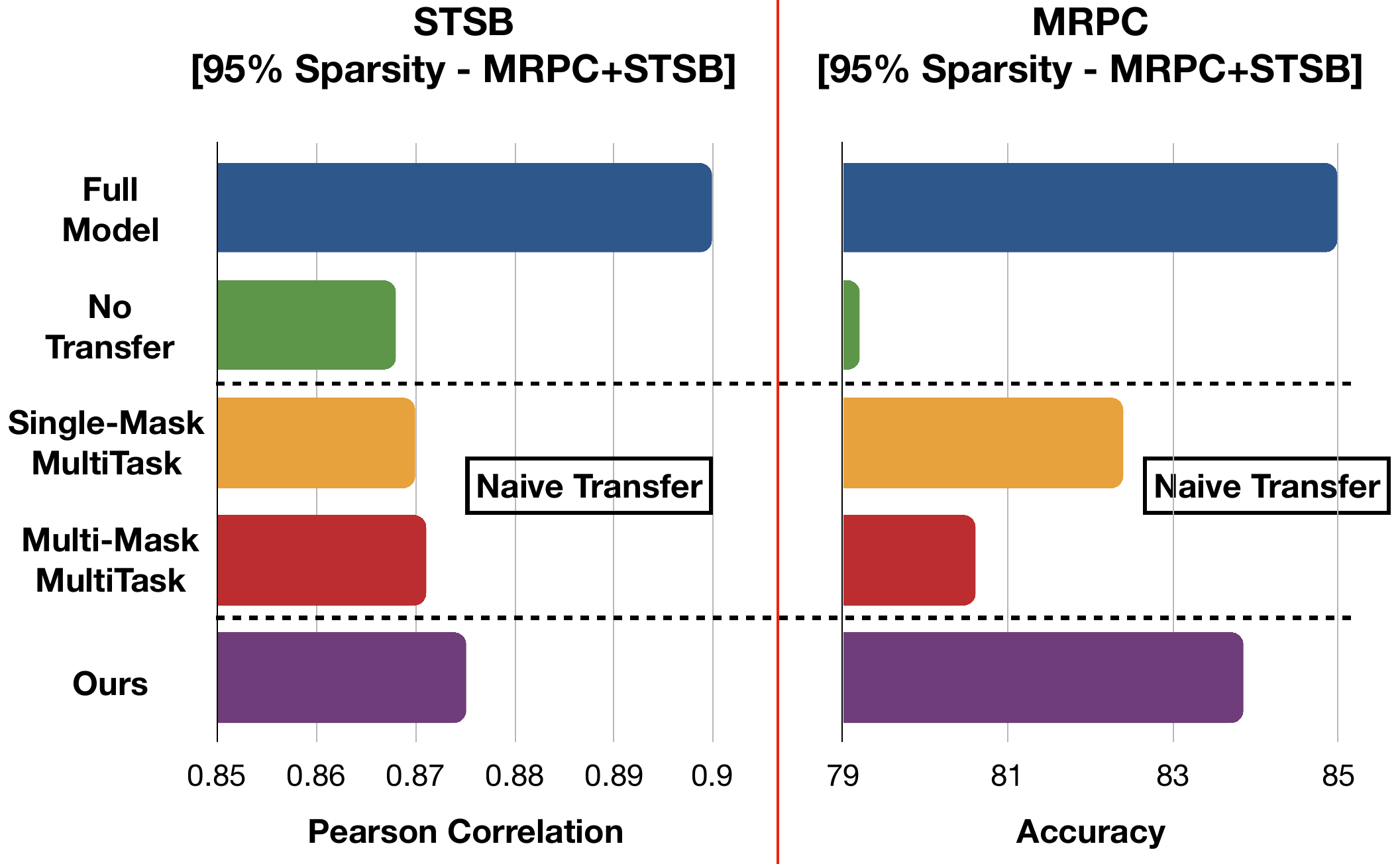}
  \end{center}
  \caption{\label{fig:joined_pairwise_stsb_mrpc} STSB and MRPC performance at 95\% sparsity. Our proposed $\delta$-Formulation outperforms all other methods on both tasks.}
\end{figure}
In Section \ref{subsec:transfer_how}, we posed different design questions around how to perform transfer learning for structured pruning under limited data and presented different options for resolving said questions. In this section, we proceed to perform a sequence of experiments to validate which choices lead to superior end-task generalization after pruning, so we can make principled recommendations to practitioners.
\subsection{How should you transfer?}
\begin{figure}
  \begin{center}
    \includegraphics[scale=0.22]{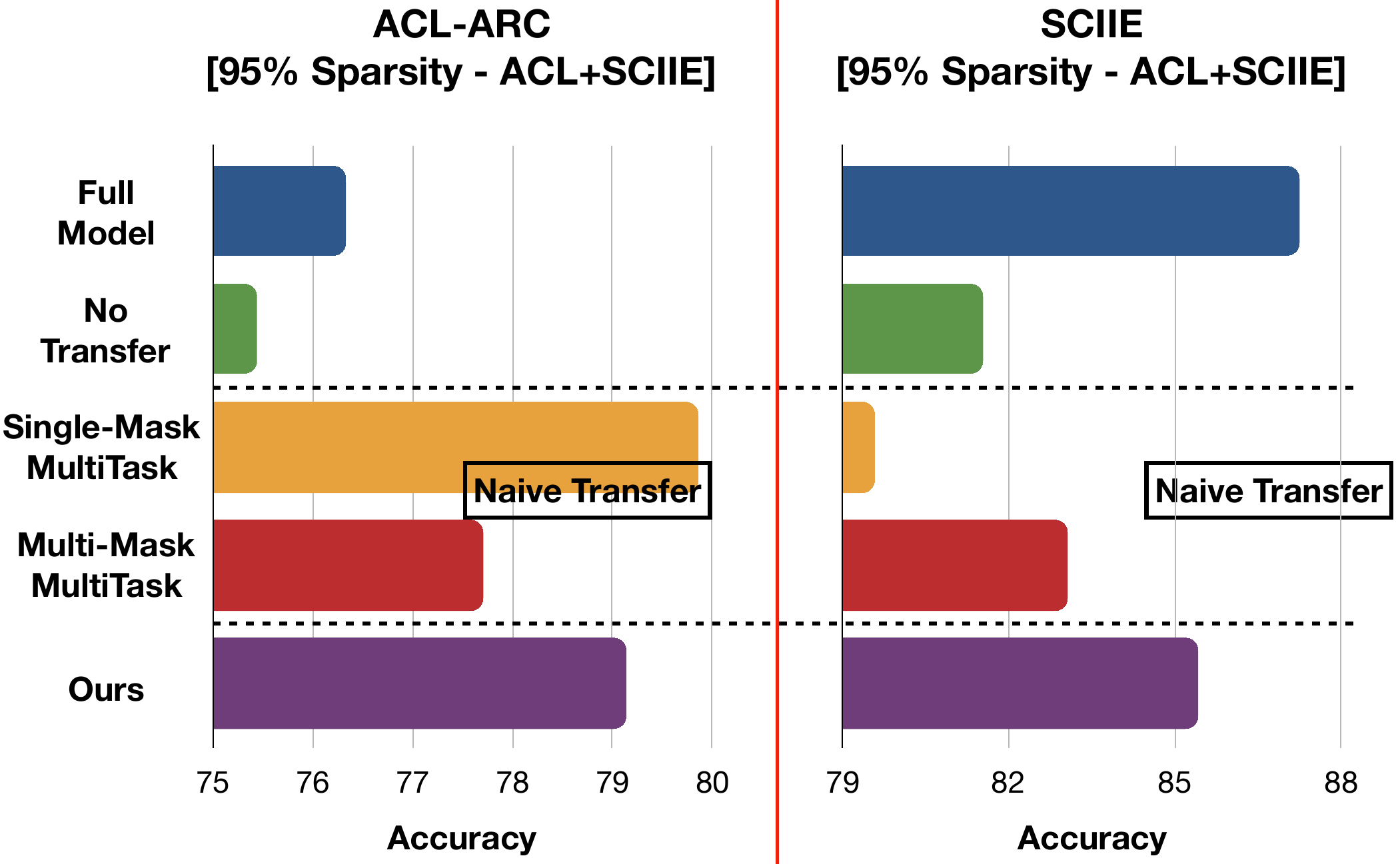}
  \end{center}
  \caption{\label{fig:joined_pairwise_sciie_acl}SCIIE and ACL\_ARC performance at 95\% sparsity. Our $\delta$-Formulation produces the best average performance across the two tasks when either is used as the auxiliary task for the other.}
\end{figure}
\begin{figure}
  \begin{center}
    \includegraphics[scale=0.22]{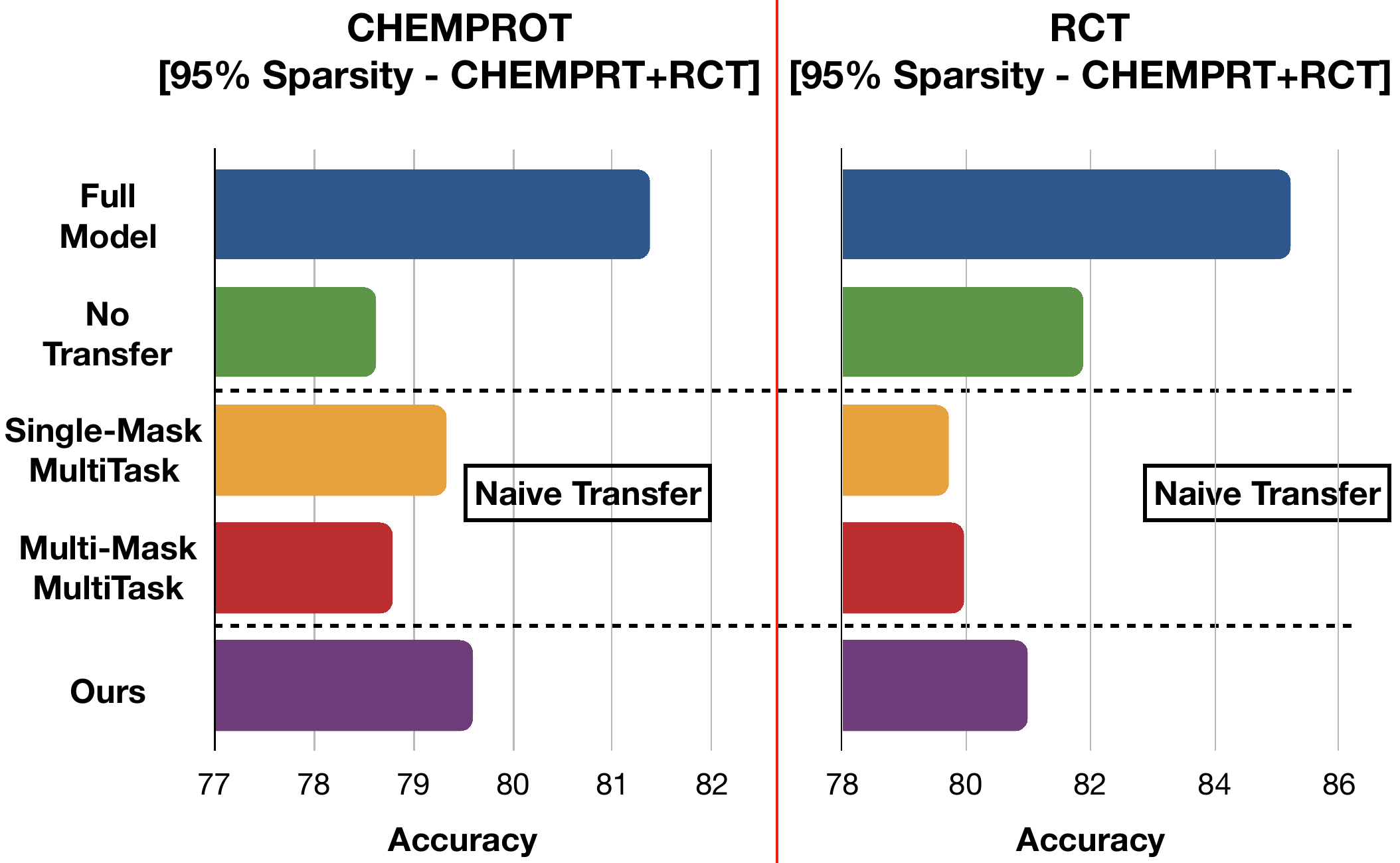}
  \end{center}
  \caption{\label{fig:joined_pairwise_rct_chemprot} RCT and Chemprot performance at 95\% sparsity. We see negative transfer from Chemprot to RCT across all transfer methods. Our proposed approach suffers least from performance degradation.}
\end{figure}

In Section \ref{subsubsec:how_learn_joint}, we introduced various approaches for coupling the auxiliary task with the target task during structured pruning. Figures \ref{fig:joined_pairwise_stsb_mrpc}, \ref{fig:joined_pairwise_sciie_acl} and \ref{fig:joined_pairwise_rct_chemprot} , show experimental results after implementing various options with different pairs of datasets. Across all dataset pairs, our $\delta$-Formulation produces the best performance when averaged across the task pair. 

For the SCIIE task, tightly coupling its structural variable with those of ACL\_ARC (as an auxiliary task) under the single-mask multi-task approach can negatively impact performance compared to not doing transfer learning at all (Figure \ref{fig:joined_pairwise_sciie_acl}). Our $\delta$-Formulation ensures that SCIIE actually benefits introducing transfer learning by outperforming the multi-mask multi-task approach that fully decouples the structural variables. For the ACL\_ARC task, our formulation recovers close to the best performance (single mask multi-task). Note that in principle, our formulation can mimic the single-mask multitask setting by using a high enough regularization on the $\delta$ offsets but we used a default $l_{2}$-regularization strength of $1e^{-2}$ for all experiments to exhibit robustness of our method. It is interesting to note that for the ACL\_ARC task, all transfer learning approaches at 95\% sparsity outperform training the full model on task data only. Note from Table \ref{table:dataset_specs} that ACL\_ARC is our smallest dataset. We posit that training the full, large model on this task leads to overfitting, resulting in poor generalization compared to leveraging transfer-learning at a reduced model size. 

Figure \ref{fig:joined_pairwise_rct_chemprot} presents an interesting scenario where Chemprot benefits from transfer but RCT does not. Whilst this could be due to the fact that we perform limited hyper-parameter tuning (mainly to exhibit the robustness of our method and to reflect compute constrained settings), it is encouraging to see that the $\delta$-Formulation for coupling structural masks significantly helped dampen the impact of negative transfer in the case of RCT as the target task.

\subsection{What should you transfer?}
So far, we have discussed using the auxiliary task when learning both the structural variables and parameters of the pruned model. In this section, we investigate if transferring both is needed. We perform the following ablation at 95\% sparsity to determine what is most important to transfer. For this, we assume that the \emph{the target task is not used during pruning} but is only introduced during the final fine-tuning of the smaller, pruned model.\\
\textbf{Weights Only}: We learn model weights and structural mask for the auxiliary task only. We then generate a random structural mask at the appropriate sparsity level (95\%) and extract the model weights corresponding to this mask from the model trained on the transfer task. We then fine-tune this smaller, pruned model on the target task. 
\\\noindent
\textbf{Masks Only}: We learn model weights and structural mask for the auxiliary/transfer task only. We then reset the model weights to the pre-trained (not-yet-finetuned) state. Given the learned mask from the transfer task, and the untuned model weights, we then fine-tune this pruned model on the target task.
\\\noindent
\textbf{Masks and Weights}: We use the transfer task to learn both the model weights and structural mask. We take weights and masks of this small model and fine-tune it on the target task.

Table \ref{table:what_to_transfer} shows the results of this ablation. For both STSB $\rightarrow$ MRPC and MRPC $\rightarrow$ STSB, we see that if we are only introducing the target task in the fine-tuning stage, it is beneficial to transfer both the weights and structure that are learned from the auxiliary task. 
\subsection{When should you transfer?}
Table \ref{table:when_to_transfer} shows results for the different choices presented in Section \ref{subsec:when_introduce_transfer} relating to when to introduce the transfer task. These experiments are also conducted at a target sparsity of 95\%. 

We obtain the best performance with the Prune$(A, T) \rightarrow $ FT$(T)$ and Prune$(A, T) \rightarrow $ FT$(A, T)$ approaches. This matches intuition because we expect an appropriately chosen auxiliary task to be helpful in terms of learning both structure and parameters of the final pruned model. Thus, introducing it in the first (pruning) stage mitigates the challenge that is exacerbated by learning a larger set of variables from limited data. 
\begin{table*}[t!]
    \begin{center}
        \caption{Transferring structure, weights or both on STSB and MRPC?  It is most beneficial to transfer both the learned weights and structural variables (masks)}
        \begin{tabular}{c c rrrr}
            \toprule
              & Metric & No Transfer &  Weights Only &  Structure Only & \multicolumn{1}{l}{Both} \\
            \midrule
             STSB $\rightarrow$ MRPC & Accuracy \%& 79.2 & 68.4 $(\downarrow)$ &  76.96 $(\downarrow)$ & 79.7 $(\uparrow)$\\
             \midrule 
             MRPC $\rightarrow$ STSB  & Pearson C.& 0.868 & 0.23 $(\downarrow)$ & 0.8527 $(\downarrow)$ & 0.871 $(\uparrow)$\\
            \bottomrule
        \end{tabular}
        \label{table:what_to_transfer}
    \end{center}
\end{table*}
\begin{table*}[t!]
    \begin{center}
    \caption{When to introduce each task on MRPC and STSB? We find that it is optimal to prune with both the auxiliary and target task jointly.}
    \resizebox{0.98\textwidth}{!}{
        \begin{tabular}{c c rrrrr}
            \toprule
             & Metric & No Transfer &  Prune$(A)$  &  Prune$(T)$  &   Prune$(T, A)$ & Prune$(T, A)$\\
             &        &             &   $\rightarrow $FT$(T)$ &  $\rightarrow 
 $FT$(A, T)$ &   $\rightarrow $FT$(T)$ &  $\rightarrow 
 $FT$(A, T)$  \\
            \midrule
             STSB $\rightarrow$ MRPC & Accuracy \%& 79.2 & 79.7 $(\uparrow)$ &  83.09 $(\uparrow)$ & 83.82 $(\uparrow)$ & 84.56 $(\uparrow)$\\
             \midrule 
             MRPC $\rightarrow$ STSB  & Pearson C.&  0.868 & 0.871 $(\uparrow)$ & 0.861 $(\downarrow)$ & 0.8751 $(\uparrow)$ & 0.872 $(\uparrow)$\\
            \bottomrule
        \end{tabular}  
        }
        \label{table:when_to_transfer}
    \end{center}
\end{table*}
\subsection{How should we choose the transfer task?}
Table \ref{table:impact_quality} contains experimental results highlighting our investigation of different variables that can impact the quality of a transfer task. 

We get the best improvements when we use a high resource auxiliary task from the same domain as the target task. As mentioned in Section \ref{subsec:how_select_task}, we use domain as a proxy for task relatedness. We see that even using a high resource task that is out-of-domain with respect to the end-task (RCT) can improve generalization over not introducing a transfer task at all ($85.29$ versus $79.2$ for MRPC) and ($0.873$ versus $0.863$ for STSB).
\begin{table*}[h!]
    \begin{small}
    \begin{center}
        \caption{Selecting the auxiliary task. A high-resource, in-domain task leads to the best result. For all experiments, best results from hyper-parameter search are reported. All models (except Full BERT) are pruned to 95\% sparsity.}
        \begin{tabular}{ccc|cccc}
            \toprule
             Target              & Full BERT &  No Transfer &  Domain &   Resourced-ness & Transfer Task & Performance\\
            \midrule
             &    &       &  In-Domain & High (364k) & QQP & $\boldsymbol{85.78}$ \\
              MRPC  &  83.48        &   79.2        &  In-Domain & Low (7k) & STSB & 83.82 \\
              &      &         &  Out-of-Domain & High (180k) & RCT & 85.29 \\
             \midrule 
              &    &       &  In-Domain & High (364k) & QQP & $\boldsymbol{0.877}$ \\
             STSB &  0.901        &   0.868    &  In-Domain & Low (3.7k) & MRPC & 0.875 \\
              &      &         &  Out-of-Domain & High (180k) & RCT & 0.873 \\
            \bottomrule
        \end{tabular}
        \label{table:impact_quality}
    \end{center}
    \end{small}
    \vspace{-5mm}
\end{table*}
\subsection{Does the learned structured sparsity translate to hardware speedups?}
So far, we have only discussed the impact of transfer learning on generalization with respect to the end-task. However, when generating pruned models, we not only care about their generalization but also the degree of speedup that is achieved at the target sparsity. 

Taking SCIIE as our primary task and ACL-ARC as the transfer task, we explore the accuracy-speedup tradeoff that is induced by leveraging transfer learning for structured pruning. We vary the degree of compression from $40\%$ sparsity to $98\%$. 
To benchmark speed, we use the wall-clock time required to perform inference on the full SCIIE dataset through the model using at a batch-size of 128. All experiments were conducted on NVIDIA V100 GPUs.
\begin{figure}
  \begin{center}
    \includegraphics[scale=0.312]{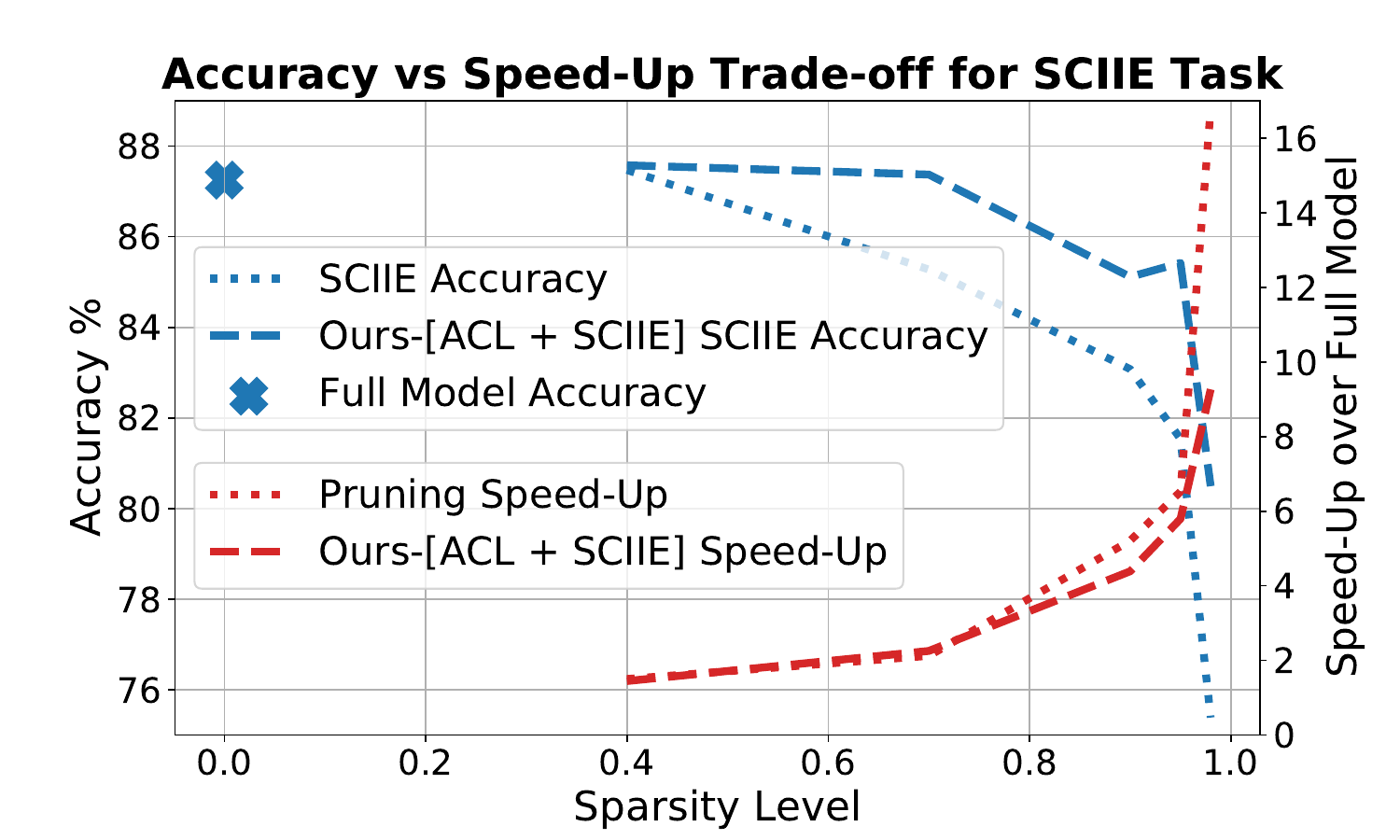}
  \end{center}
  \caption{\label{fig:acc_vs_speed_tradeoff}Accuracy vs speed tradeoff on SCIIE. Our $\delta$-formulation, gives accuracy boosts on SCIIE at varying levels of compression.}
\end{figure}
Figure \ref{fig:acc_vs_speed_tradeoff} summarises our findings. For SCIIE+ACL, we fix the task weighting to the best performing configuration from our 95\% sparsity experiments, the rest of the hyper-parameters are cross-validated from values in Table \ref{tab:hp_choices}.
At 50$\times$ compression (95\%) sparsity, we are able to obtain a $\sim5\%$ boost in accuracy over not using a transfer task, whilst achieving a $\sim10\times$ speedup in inference. Transfer leraning enables a more graceful degradation in accuracy (dashed blue line) whilst still finding pruned models with comparable speed-ups.

Another view of Figure \ref{fig:acc_vs_speed_tradeoff} is to consider the model size required for a threshold level of accuracy for deployment. At a threshold of 84\% accuracy, whilst naive pruning results would produce a model at 80\% sparsity, we are able to produce one at 95\% sparsity! This is a $\sim 1.2\times$ memory saving and $\sim 2.8\times$ inference speedup.
\subsection{What are the structural differences between a pruned model using transfer learning and without?}
\begin{figure}[h!]
    \centering
    \includegraphics[scale=0.34]{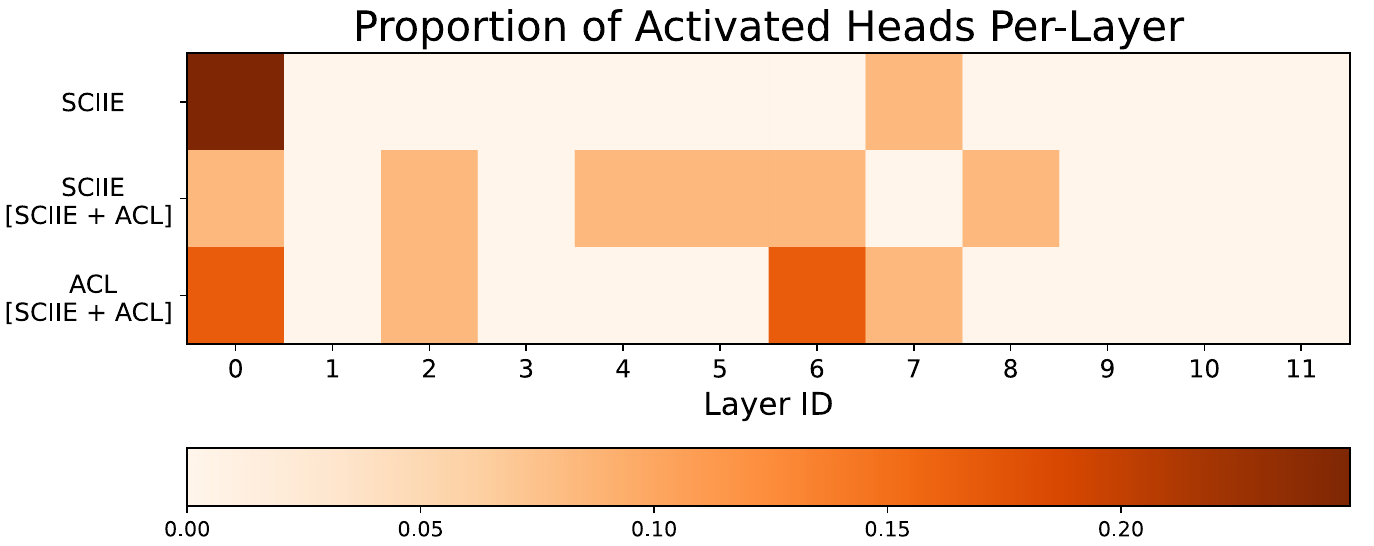}
    \caption{\label{fig:attn_heat_map} Structural visualization at 98\% sparsity. Qualitatively, using a transfer task changes the pruned model structure significantly. The ACL transfer task in this case induces the learned SCIIE structure to be more diffuse across the layers of the model.}
\end{figure}
At extreme sparsity levels, the differences in speedup from learning a pruned model with and without transfer learning (Figure \ref{fig:acc_vs_speed_tradeoff}) suggest that the models discovered have different structures. 

\begin{figure}[h!]
    \centering
    \includegraphics[scale=0.34]{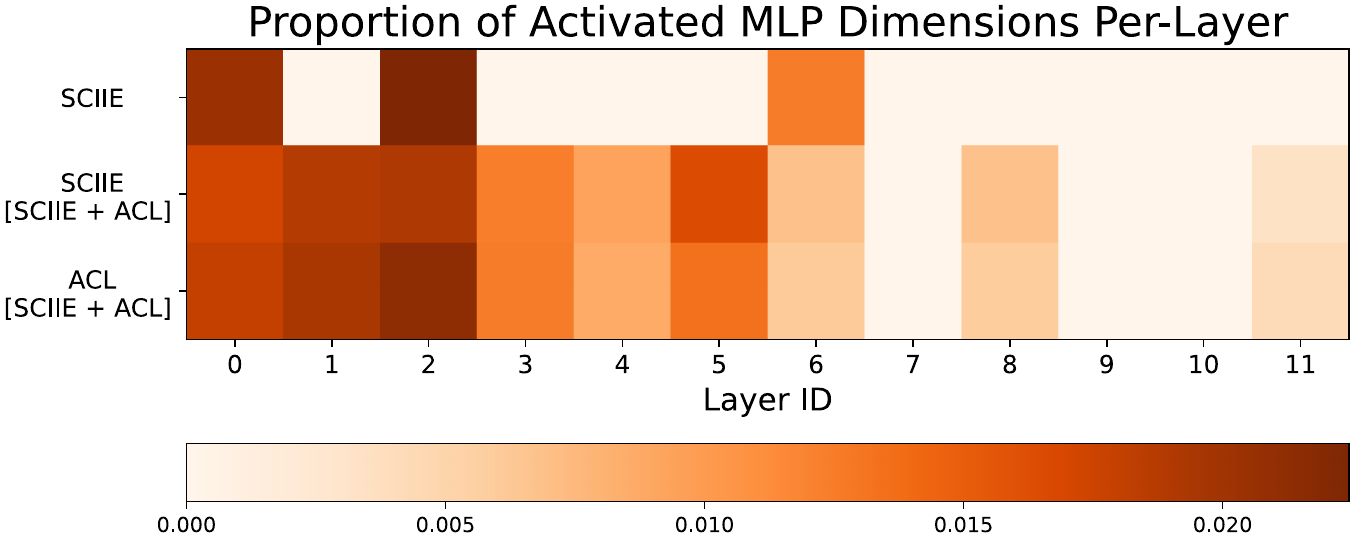}
    \caption{\label{fig:mlp_heat_map}  Structural visualization at 98\% sparsity. Qualitatively, using a transfer task changes the pruned model structure significantly. The ACL transfer task in this case induces the learned SCIIE structure to be more diffuse across the layers of the model.}
    
\end{figure}

Figures \ref{fig:attn_heat_map} and \ref{fig:mlp_heat_map} show the fraction of attention heads and MLP intermediate dimensions respectively, that are preserved across each layer with respect to the original BERT$_{\mathrm{base}}$ model. Pruning with the target task alone results most of the preserved parameters coming from earlier in the network. With an auxiliary task however, the pattern of preserved modules is more diffuse across layers. We posit that this results from the the two tasks being multi-tasked preferring different layers thus resulting in a more diffuse distribution of preserved modules as a compromise in order to perform reasonably well on both tasks.
\section{Conclusion}
As coined in \citet{ahia2021low}, the \emph{low-resource double bind} describes the challenge of producing compressed models to serve compute-starved (memory and latency limits) tasks under a setting where these tasks also have limited data for pruning.  In this work, we have explored adapting transfer learning, which has traditionally been leveraged only for learning model weights, to robustly prune models when the target task is data-limited.

We have provided practitioners with recommendations on how to choose a transfer task, when and how to incorporate it into the pruning optimization procedure and what elements to transfer from the auxiliary task to the target. Equipped with this knowledge, we plan to explore the problem of structured pruning under limited target data for larger scale models.

%% file: sections/appendix.tex
\appendix
\section{Datasets}

Table~\ref{table:dataset_specs} describes the tasks and datasets used in our experiments.
\label{app:data_info}
\begin{table*}[t!]
    \begin{small}
    \begin{center}
        \caption{Specifications of datasets used to evaluate our methods. \\}
        \label{table:dataset_specs}
        \begin{tabular}{llllllll}
            \toprule
             Domain & Task & Task-Type & Train Size & Metric \\
            \midrule
             BIOMED & CHEMPROT   \cite{kringelum2016chemprot} & Classification & 4169 & Accuracy\\
              & RCT   \cite{dernoncourt2017pubmed} &  Classification & 10K$^{*}$ & Accuracy\\
            \midrule
             CS & SCIIE   \cite{luan2018multi}   & Classification & 3219 & Accuracy \\
             & ACL-ARC   \cite{jurgens2018measuring}   & Classification & 1688 &  Accuracy\\
            \midrule
            GLUE & STSB   \cite{wang2018glue}   & Sentence Similarity & 7K & Pearson's Correlation \\
             & MRPC   \cite{wang2018glue}   & Paraphrase Detection & 3.7K &  Accuracy\\
            \bottomrule
        \end{tabular}
    \end{center}
    \end{small}
\end{table*}

\section{Training Details} 
\label{app:training_details}
We follow as closely as possible the hyper-parameters that are used in the original CoFi code base. 
Table~\ref{tab:hp_choices} reports CoFi-specific hyper-parameter settings. 

Unlike the original CoFi, we turn off output prediction distillation for all experiments. ie -- we do not distill the predictions from the pre-trained models since unlike in the original CoFi paper, we are not starting from a model that has already been fine-tuned on the target task but rather we are starting from the pre-trained model itself.

During pruning, we perform 10K gradient descent steps to learn both the structural and parameter variables 
of the model. We perform 20 epochs of post-pruning finetuning on the target task.

\begin{table}[h!]
    \centering
    \begin{small}
    \caption{Hyper-parameter choices}
    \label{tab:hp_choices}
    \begin{tabular}{lll}
    \toprule
          Hyper-parameter & Values & Description \\
    \midrule
        Task pair weightings & (1, 1), (1, 2), (2, 1) & Weightings applied to transfer task vs target task during training.\\
        Model LR - Pruning & 1e-4, 2e-5 & Learning rate used for model parameters during pruning. \\
        Model LR - Finetuning & 1e-4, 2e-5 & Learning rate used for finetuning pruned model. \\
        Structure LR & 0.1, 0.01 & Learning rate used for learning structural parameters. \\
        $\delta$-$l_{2}$ Reg Weight & 1e-2 &  Regularization weight used in  $\delta$-formulation. \\
    \midrule
    \end{tabular}
    \end{small}
\end{table}